\documentclass[fleqn,10pt]{wlscirep}
\usepackage[utf8]{inputenc}
\usepackage[T1]{fontenc}
\usepackage{hyperref}
\usepackage{color}

\title{Differentiable multiphase flow model for physics-informed machine learning in reservoir pressure management}

\author[1,*]{Harun Ur Rashid}
\author[1]{Aleksandra Pachalieva}
\author[1]{Daniel O'Malley}
\affil[1]{Earth and Environmental Sciences Division, Los Alamos National Laboratory, Los Alamos, NM, 87545, USA}
\affil[*]{Corresponding author(s). E-mail(s): hrashid@lanl.gov }
\usepackage{listings}
\usepackage{xcolor}
\usepackage{graphicx}
\usepackage{subcaption}
\begin{abstract}

Accurate subsurface reservoir pressure control is extremely challenging due to geological heterogeneity and multiphase fluid-flow dynamics. Predicting behavior in this setting relies on high-fidelity physics-based simulations that are computationally expensive. Yet, the uncertain, heterogeneous properties that control these flows make it necessary to perform many of these expensive simulations, which is often prohibitive. To address these challenges, we introduce a physics-informed machine learning workflow that couples a fully differentiable multiphase flow simulator, which is implemented in the DPFEHM framework with a convolutional neural network (CNN). The CNN learns to predict fluid extraction rates from heterogeneous permeability fields to enforce pressure limits at critical reservoir locations. By incorporating transient multiphase flow physics into the training process, our method enables more practical and accurate predictions for realistic injection-extraction scenarios compare to previous works. To speed up training, we pretrain the model on single-phase, steady-state simulations and then fine-tune it on full multiphase scenarios, which dramatically reduces the computational cost. We demonstrate that high-accuracy training can be achieved with fewer than three thousand full-physics multiphase flow simulations -- compared to previous estimates requiring up to ten million.This drastic reduction in the number of simulations is achieved by leveraging transfer learning from much less expensive single phase simulations.
\end{abstract}

\begin{document}

\flushbottom
\maketitle

\thispagestyle{empty}
\section*{Introduction}
Reservoir pressure management refers to the control and maintenance of fluid pressure within a subsurface formation to optimize injection or extraction and prevent adverse effects caused by excessive or insufficient pressure. This is typically achieved through the injection or extraction of fluids to maintain a target pressure level within the reservoir~\cite{archer1986secondary,jansen2009closed,jiao2017field}. The importance and application of pressure management span various fields, including carbon sequestration~\cite{lal2008carbon,figueroa2008advances,simmenes2013importance,boot2014carbon}, oil and gas production, subsurface gas storage~\cite{birkholzer2012impact,liu2015storage}, geothermal energy extraction~\cite{ungemach2005sustainable,acuna2008reservoir,ghassemi2012review}, and wastewater disposal~\cite{fragachan2006pressure,barbour2019leakage}.

Reservoir pressure management is essential since uncontrolled pressure can trigger seismicity and rock fracturing. Induced seismicity poses significant environmental and operational risks, including the potential for damaging earthquakes~\cite{majer2007induced,rashid2024use}. For example, large-scale wastewater re-injection in central Oklahoma led to unexpected seismic events~\cite{zoback2012managing, keranen2014sharp, mcnamara2015earthquake}, while high-pressure water injection into the Basel geothermal reservoir in Switzerland induced seismicity that ultimately forced the cancellation of the project~\cite{baer2007earthquakes, deichmann2008earthquakes, dyer2008microseismic}. Preventing induced seismicity is also critical in subsurface gas storage (e.g., CO$_2$ and hydrogen), where long-term caprock integrity must be maintained to prevent leakage, which poses both environmental and economic risks~\cite{buscheck2011combining, cihan2015optimal, harp2017development}.

Most subsurface reservoirs are highly heterogeneous and uncertain. Their properties are often poorly characterized, fluids exhibit complex multiphase behavior, and the domains are typically large in scale~\cite{alabert1990heterogeneity,tavakoli2019carbonate, khalili2023reservoir}. A reliable pressure management tool must address these challenges and provide rapid predictions. However, the complexity of subsurface physics makes full-physics modeling expensive, and it is difficult to design a strategy that is both accurate and efficient enough for real-time decision-making.

Many high-fidelity simulation tools are available to predict subsurface pressure when reservoir properties are known~\cite{pruess1991tough2,pettersen2006basics, dai2014integrated, rashid2024continuous}. However, designing a robust pressure management strategy requires simulations of heterogeneous reservoir properties and multiphase fluid behaviors across a wide range of geological scenarios. In practice, this means running high-fidelity physics-based models on thousands of permeability realizations or more to account for the uncertainty and risk inherent to these problems. Unfortunately, traditional reservoir simulators struggle to handle the combined challenges of reservoir heterogeneity, multiphase flow, and uncertainty, driving computational cost so high that routine pressure modeling accounting for all these factors becomes untenable. 

To over come the high computational cost, one potential solution is the use of surrogate machine learning (ML) models, which promise orders-of-magnitude speedups over physics-based simulators. While numerous ML studies in the reservoir engineering literature have addressed CO$_2$ storage~\cite{chen2018geologic, menad2019predicting, sinha2020normal, wang2020inferring, yan2022gradient, liu2024prediction, alqahtani2023uncertainty, chu2022deep}, enhanced oil recovery~\cite{cheraghi2021application, pirizadeh2021new, krasnov2017machine, you2019assessment, kumar2023supervised, mahdaviara2022toward, vaziri2024machine, khan2024application}, and geothermal energy~\cite{li2017machine, rezvanbehbahani2017predicting, holtzman2018machine, tut2020prediction, ahmmed2022machine, feng2025geothermal, yan2023robust, he2022machine}, few studies target ML applications specifically for reservoir pressure management. Despite its critical role in preventing induced seismicity and caprock failure, ML in pressure reservoir management remains underdeveloped, highlighting the need for approaches that can learn effectively with sparse data while respecting underlying physics. Moreover, these models typically depend on large, high-quality datasets, such as well logs data and production histories, which are not readily available for many storage and disposal projects. When large datasets are available, as is often the case in the oil and gas industry, the data-driven models tend to perform well. However, for CO2 storage and similar applications, acquiring sufficient data is difficult, making it challenging to train accurate models. In addition, purely data-driven models often struggle to capture complex reservoir behavior, such as multiphase interactions and geological heterogeneity, because they typically lack integration with underlying physical laws~\cite{lim2005reservoir, singh2007neural, gaganis2012machine}. Trying to optimize pressures using a data-driven model is also prone to the sort of adversarial examples that plage many data-driven models.

To overcome the physics blindness in data driven models, several studies have embedded physical constraints (e.g. mass and momentum conservation) directly into the loss function of the neural network (NN), known as physics-informed neural networks (PINNs)~\cite{raissi2019physics, yang2019adversarial, meng2020ppinn, yan2024physics, ishitsuka2023physics, wang2024hybrid}. PINNs are often insufficient in complex reservoir settings, resulting in overly simplified or approximate physics constraints. PINNs only softly constrain to the physics via the loss function and gradient descent, whereas traditional physics simulators have hard constraints that ensure the equations are accurately solved. The soft constraints often cause models to converge to incorrect solutions and produce flawed pressure management strategies.

A promising alternative to the limited physics constrained surrogate models is embedding a full-physics numerical simulator directly into the machine learning training loop, which promises more accurate, physics-consisted learning. This approach requires the simulator to be fully differentiable. However, traditional numerical models rely on finite-difference gradient estimates, which are incompatible with backpropagation algorithms used in gradient-based optimization~\cite{pruess1991tough2, pettersen2006basics, rashid2022iteratively}. To address this challenge, simulators can be developed using differentiable programming~\cite{innes2019differentiable} and automatic differentiation, enabling efficient and accurate gradient computation using the chain rule~\cite{baydin2018automatic}. Differentiable programming enables the backpropagation that has traditionally been used with NNs to be fused with the adjoint methods that have been traditionally used in solving differential equations. Differentiable programming structures the simulator explicitly for integration into ML workflows and can be implemented using computational frameworks such as PyTorch ~\cite{ketkar2021introduction} and Julia ~\cite{innes2019differentiable}.

Despite recent advances in differentiable reservoir simulators~\cite{o2023dpfehm, moyner2024jutuldarcy}, their application to ML-driven pressure management problems remains limited. Pachalieva et al.~\cite{pachalieva2022physics} recently demonstrated the use of a fully differentiable single-phase steady-state flow model within a physics-informed ML framework, but to date no work has tackled multiphase flow models. Given that multiphase behavior dominates most reservoir operations~\cite{wu2015multiphase,rashid2024continuous}, integrating a differentiable transient multiphase simulator into the ML training loop is essential for enhancing accuracy in real-world pressure-control scenarios.

One of the major challenges of incorporating a multiphase flow simulator into a machine learning workflow is that many gradient descent steps, and if implemented in the obvious way, a proportional number of multiphase simulations would need to be performed.  Harp et al.~\cite{harp2021feasibility} identified that tens of millions of multiphase simulations would be required, which would be infeasible. Here, we circumvent this limitation by using transfer learning to dramatically reduce the computational cost of the training.

In this study, we introduce an ML workflow designed to determine the extraction rate required to maintain a prescribed pressure at critical locations in the reservoir during the injection period. Our training data consists of an ensemble of heterogeneous permeability fields, and we employ a transfer learning strategy, where we pretrain the model using a fully differentiable steady-state single-phase simulator and then finetune it to transient multiphase flow simulator. The key contributions of our work are (1) embedding a fully differentiable transient multiphase flow simulator in an ML training workflow; (2) employing transfer-learning strategies by pretraining on single-phase steady-state models before fine-tuning on multiphase flows.

This work is organized as follows. In the Method section, we review the background physics and present the training workflow for reduced-order modeling. The Results section presents training and validation outcomes and demonstrates model accuracy through a case study and statistical analysis. In the Discussion section, we highlight the strengths and limitations of the proposed pressure management workflow and outline directions for future improvement. Finally, in the Conclusions, we summarize the novelty, strengths, applicability, and limitations of our approach.

\section*{Methods} 
The primary objective of this study is to develop a surrogate model capable of determining the optimal fluid extraction rate required to manage overpressure within a subsurface reservoir. To achieve this, we train a NN using heterogeneous permeability fields as input to predict the optimal extraction rate. The predicted extraction rate is then passed through a full-physics, differentiable simulator, which evaluates the resulting pressure at a designated critical monitoring location. The simulated pressure is compared against a prescribed target value, and the resulting difference is used to compute the loss function for training the surrogate model.

In this section, we outline the complete methodology employed in this study. We begin by presenting the governing equations that underpin the physics-based simulator. This is followed by a description of the problem setup and the numerical solver used to compute the physical response of the system. We then illustrate the architecture of the NN model. Finally, we describe the training workflow in two stages: initially, the training loop incorporates only the multiphase simulator; subsequently, a transfer learning strategy is introduced, which uses a steady-state single-phase model in combination with a transient multiphase solver to accelerate training efficiency.

\subsection*{Physics model} 
The full-physics simulator, central to our training loop, is governed by a set of flow equations designed to capture both steady-state single-phase and transient multiphase dynamics. Since our workflow employs a transfer learning approach that integrates both steady-state single-phase flow and transient multiphase flow, we provide a detailed description of the flow equations, their numerical discretizations, and the corresponding simulators for each set of equations. Both physics-based simulators are available as open-source tools within the DPFEHM GitHub repository~\cite{o2023dpfehm}.

In the pretraining stage, we use a single-phase steady-state flow model, which assumes that pressure changes resulting from injection or extraction occur in a single-phase fluid and are independent of time~\cite{von1956mechanics}. In heterogeneous reservoir, this leads to solving the following partial differential equation:
\begin{equation}
\nabla \cdot \left( K(x) \cdot \nabla p \right) = q,
\end{equation}
where $p$ is the pressure, $K(x)$ is the spatially varying permeability field, and $q$ represents sources and sinks. The use of the steady-state single-phase flow equation enables rapid evaluation of how different extraction rates influence the pressure field, making it particularly well-suited for efficient pretraining.

For the final training, the pressure change in the reservoir is calculated by considering transient multiphase fluid flow through a heterogeneous permeability field. Such a model involves solving the mass conservation equations for the phases present in the flow. Our differentiable simulator uses a sequential implicit pressure and explicit saturation (IMPES) scheme for incompressible and immiscible two-phase flow, as described by Hasle et al.~\cite{hasle2007geometric}. As in the Hasle et al.~\cite{hasle2007geometric}, we neglect gravity and capillary forces for simplicity. Based on these assumptions, we solve the following governing equations for pressure ($p$) and saturation ($s$) respectively:
\begin{align}
    &\text{Pressure equation:}\quad\;\; -\nabla\cdot K(x)\lambda(s)\nabla p = q, 
    \label{eq:pressure_eq}\\
    &\text{Saturation equation:}\quad\phi \frac{\partial s}{\partial t}+\nabla \cdot \left(f(s)v\right) = \frac{q_w}{\rho_w}.
    \label{eq:saturation_eq}
\end{align}
Here, $\lambda$ in pressure equation is the mobility of a phase and a function of saturation. In the saturation equation the term $\phi$ represents the porosity of the medium, $t$ is time , $f(s)v$ represents the viscous force, where $v$ is the velocity of the phase and $f(s)$ is fractional flow function. On the right side of the saturation equation, $q_w$ is the wetting phase injection rate and $\rho_w$ represents the density of wetting phase. The term $\frac{q_w}{\rho_w}$  represents the source. In our simulator we only inject the wetting phase. Therefore the source term can be modifies as follows: 
\begin{equation}
    \frac{q_w}{\rho_w}=max(q,0)+f(s)min (q,0).
\end{equation}
In our model, we calculate the mobility of wetting and non-wetting phase using the following analytical expression:
\begin{equation}
    \lambda_w(s)=\frac{(s^*)^2}{\mu_w}.
\end{equation}
\begin{equation}
    \lambda_{nw}(s)=\frac{(1-s^*)^2}{\mu_o},
\end{equation}
with $s^*$ defined as follows:
\begin{equation}
    s^*=\frac{s-s_{wc}}{1-s_{nwr}-s_{wc}}.
\end{equation}
Here, the term $s_{nwr}$ represents the irreducible saturation for the non-wetting phase, which is the minimum amount of fluid that remain in the reservoir despite the displacement, and $s_{wc}$ is the connate saturation saturation for the wetting phase, which is the minimum fluid saturation exist naturally in the pores and generally immobile due to high capillary pressure.  

Finally we can discretize the saturation equation using finite-volume scheme:
\begin{equation}
    s_{i}^{n+1}=s_i^n+(\delta_x^t)_{i}\left(max(q_i,0)-\sum_{j}f(s^n)_{ij}v_{ij}+f(s^{n}_{i}) min(q_{i},0)\right),
\end{equation}
where the superscript, $n+1$ represent the value of the variable in the current time step, and $n$ represents the value in the previous time step. We calculate the term $(\delta_x^t)_{i}$ based on the CFL (Courant–Friedrichs–Lewy) time step condition~\cite{courant1928partiellen}, which determines the maximum allowable timestep that ensures numerical stability and prevents non-physical oscillations or instabilities during simulation.

To solve the single-phase and multiphase flow equations, we utilize the DPFEHM framework~\cite{o2023dpfehm}. DPFEHM is an automatically differentiable Julia package that implements standard two-point flux approximation and finite volume methods for numerical discretization. Its built-in support for automatic differentiation enables seamless integration of the full-physics models into our machine learning workflow, allowing for efficient backpropagation through the simulation process.

\begin{figure}[h!]
  \centering
  \includegraphics[
    width=0.6\linewidth]{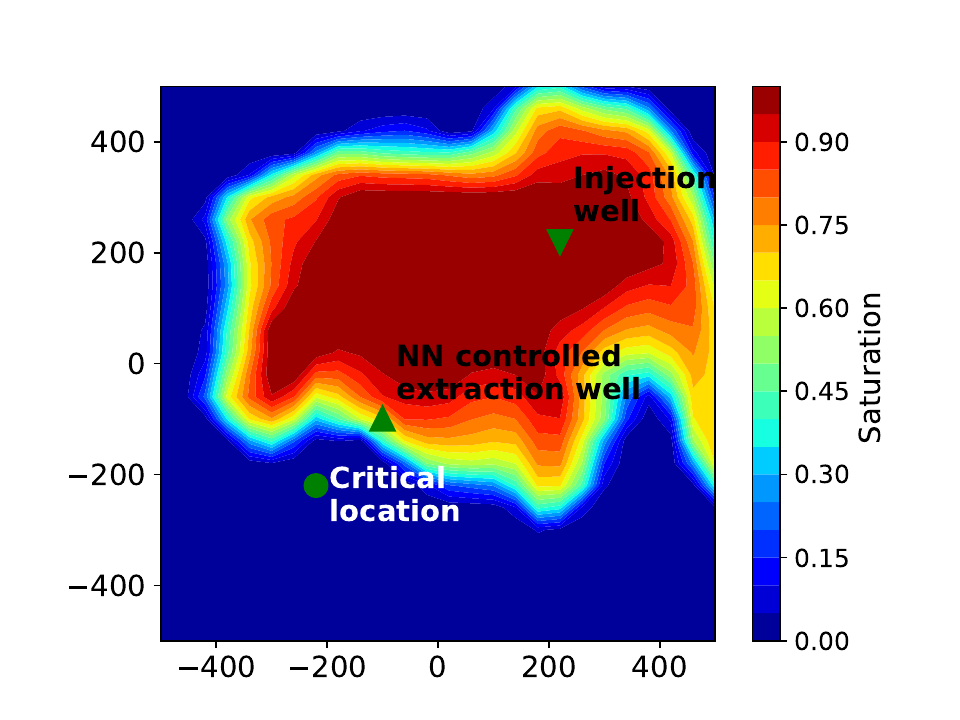}
  \caption{Simulation domain for the pressure management workflow, showing well locations and the saturation front of the injected fluid in the background. The NN–controlled extraction well withdraws a portion of the injected fluid to maintain the prescribed pressure at the critical location. }
  \label{fig:Application_pressure_setup}
\end{figure}

Figure \ref{fig:Application_pressure_setup} illustrates the layout of the simulation domain. The simulation is conducted on a 2D domain measuring 1000 by 1000 meters. The setup includes one injection and one extraction well, marked with a downward-pointing and upward-pointing triangle, respectively. Fluid is injected at a constant rate of 0.031688 $m^3/s$ (approx.\,1 Mt/year) at the injection location. The critical location is marked by a circle. Our pressure management workflow aims to maintain a prescribed pressure at this location throughout the injection period, which is ensured by the optimum extraction rate at the extraction well. The extraction well is strategically positioned between the injection and the critical locations, and is closer to the latter to maximize its control over the pressure in the target region. For this study, we used the simulation and geostatistical parameters listed in Table\ref{tab:param_physics}.

\begin{table}[h]
\begin{center}
\caption{Parameters used in physics model}\label{tab:param_physics}%
\begin{tabular}{@{}lll@{}lll@{}}
    \toprule
    Parameters & Value &  Unit\\
    \midrule
    Boundary condition    & Dirichlet pressure & -  \\
    Boundary pressure value   & 0 & Pa   \\
    Irreducible fluid saturation    & 0 & - \\
    Connate fluid saturation    & 0 & - \\
    Porosity   & 1 &- \\
    Viscosity of wetting phase   & 1  & Pa-s\\
    Viscosity of non-wetting phase  & 1 & Pa-s\\
    Variogram (Covariance) type & Matern, exponential & -\\
    Correlation length  & 1 &  m \\
    Sampling Method  & Karhunen–Loève (KL) & -  \\
    Number of KL modes  & 200 & -\\
    \bottomrule
\end{tabular}
\end{center}
\end{table}

\subsection*{Neural network model}
With the physics solver in place, we define NN architecture our surrogate model. Since the input is a two-dimensional, spatially varying permeability field, we employ a convolutional neural network (CNN), which is well-suited for processing structured spatial data. Specifically, our model builds upon the classic LeNet-5 architecture proposed by LeCun et al.~\cite{lecun1989backpropagation}, which combines convolutional encoding layers with fully connected dense layers to process spatially structured inputs. In this study, we adopt a slightly modified version of LeNet-5, closely following the physics-informed architecture introduced by Pachalieva et al.~\cite{pachalieva2022physics}.

Our CNN architecture consists of two convolutional layers, each followed by a max-pooling subsampling layer, a flattening step, and a fully connected dense block. The convolutional layers apply $5\times5$ kernels to extract spatial features from the input permeability field. The first layer outputs 6 feature maps, while the second produces 16. Each convolutional layer is followed by a max-pooling layers with a stride of 2, which reduces the spatial resolution by a factor of 4 across the two layers, thereby enhancing computational efficiency. The output of the convolutional block, originally in a four-dimensional tensor format, is flattened into a two-dimensional vector to interface with the subsequent dense block. This dense block consists of three fully connected layers containing 120, 84, and 1 neurons, respectively. Rectified Linear Unit (ReLU) activation function are applied to all hidden layers, defined as $\sigma(x) = \max(0, x)$, where $x$ is the input and $\sigma(x)$ is the activated output. The final layer yields a single scalar value, representing the predicted extraction rate. The NN parameters used for this study are listed in Table \ref{tab:param_NN}, and the modified CNN architecture is summarized below:

\begin{lstlisting}[language=Matlab, caption={}]
    model = Chain(Conv((5, 5), 1=>6, relu),
                  MaxPool((2, 2)),
                  Conv((5, 5), 6=>16, relu),
                  MaxPool((2, 2)),
                  Flux.flatten,
                  Dense(144, 120, relu),
                  Dense(120, 84, relu),
                  Dense(84, 1)) |> f64
\end{lstlisting}

\begin{table}[h]
\begin{center}
\caption{Parameters used in training NN model}\label{tab:param_NN}%
\begin{tabular}{@{}ll@{}}
    \toprule
    Parameters & Value   \\
    \midrule
    Learning rate    & $10^{-4}$   \\
    Batch size   & 20    \\
    Samples per batch    & 10  \\
    Samples per epoch    & 200  \\
    Optimizer    & ADAM  \\
    \bottomrule
\end{tabular}
\end{center}
\end{table}
The permeability samples used for training are randomly generated and vary in each epoch, thus the NN model never sees the same permeability field during training. For validation, 200 samples are fixed throughout the process, while the 10,000 test samples -- used to evaluate the model’s performance -- are also randomly generated.
\subsection*{Quantification of prediction errors}
To evaluate model performance and quantify prediction errors, we define a loss function based on the difference between  the simulated and prescribed pressures at the critical location. The loss function is given by:
\begin{equation}
    \mathcal{L}=\sum_{i}^{N_b}\sum_{j}^{N_s}\left[\Delta p \left(q_{nn}(\theta,k_j(x)),k_j(x)\right)-\Delta p^{target}\right]^2,
\end{equation}
where $N_b$ is the number of training batches, $N_s$ is the number of samples per batch, $p^{target}$ is the prescribed targeted pressure, and $\Delta p$ is the simulated pressure response based on the the predicted extraction rate, $q_{nn}(\theta,k_j(x))$ generated by the NN model. The NN model predicts the extraction rate based on model parameters $\theta$ and the permeability field $k_j(x)$. During training, the NN minimizes the loss function by updating its parameters through backpropagation. This iterative process drives the model toward an optimal estimation of the extraction rate, which minimizes pressure error at the critical location.

To evaluate the model's performance, we compute the root mean squared error (RMSE), defined as:
\begin{equation}
    RMSE(\theta)=\sqrt{\frac{\mathcal{L}(\theta)}{N_b N_s}}.
\end{equation}
The RMSE is a standard metric for evaluating model accuracy across all training samples, enabling consistent comparison and monitoring of convergence during the training process.

\subsection*{Pressure management workflow}
\begin{figure}[h!]
\centering
\includegraphics[width=0.85\linewidth]{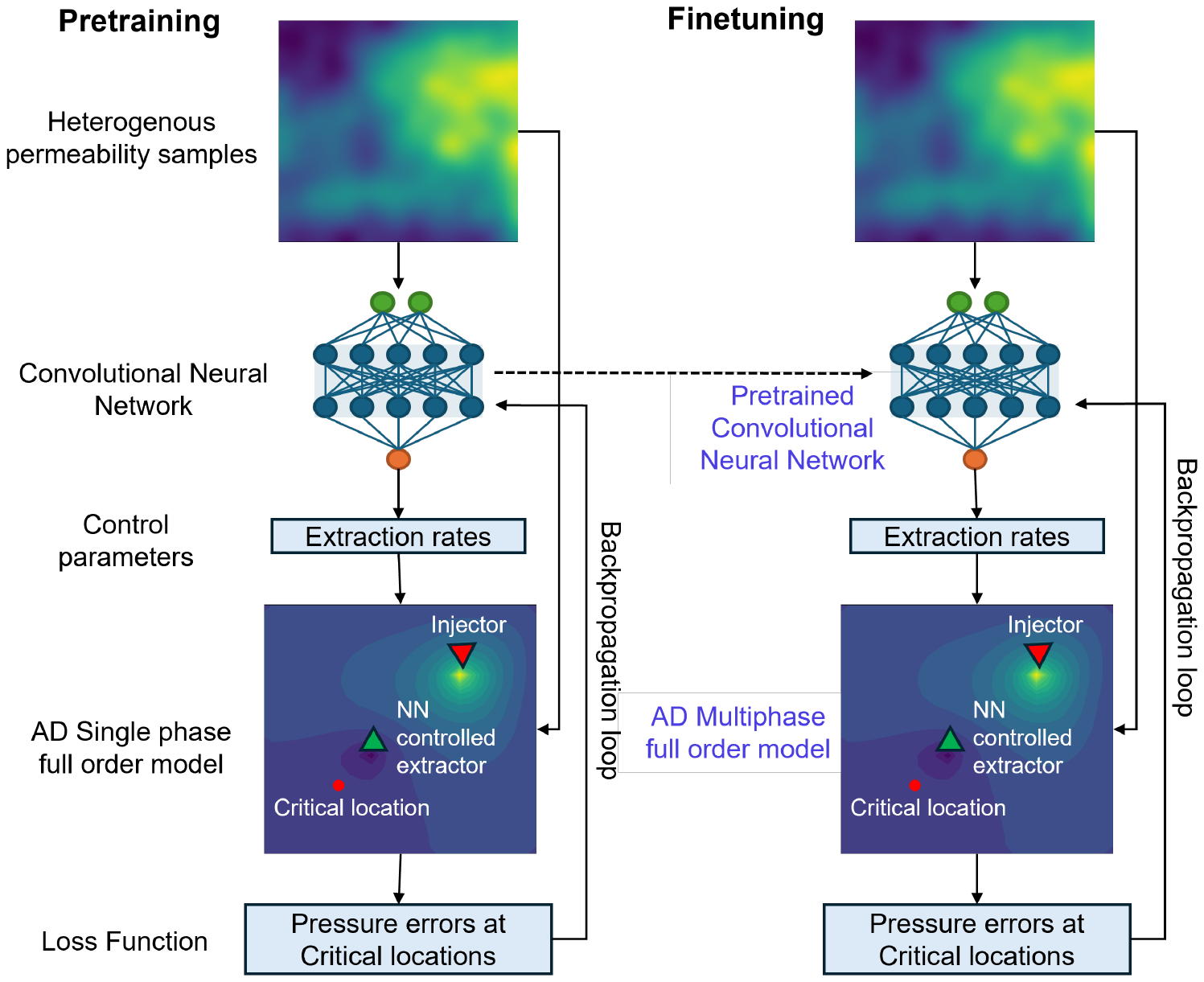}
\caption{Full training workflow for the pressure management model using a transfer learning approach, which enables efficient incorporation of complex physics. The model is initially pre-trained using fast single-phase steady-state simulations, and then fine-tuned with transient multiphase simulations during the final training stage.}
\label{fig:Application_pressure_workflow_curriculum}
\end{figure}

Having set up both the physics-based simulator and the NN model, we now integrate them into a unified training workflow designed for pressure management in the subsurface. This workflow, illustrated in Figure \ref{fig:Application_pressure_workflow_curriculum}, follows a two-stage approach consisting of pretraining and finetuning stages. Each stage begins by sampling random realizations of heterogeneous permeability fields and specifying a target pressure at a critical reservoir location. Each permeability realization are passed through the NN, which predicts the corresponding extraction rate. The predicted extraction rate is then used as input to the full-physics simulator, which computes the resulting pressure distribution. The pressure at the critical location is compared to the prescribed target, and the difference, referred to as the pressure error, is used to perform backpropagation. This error signal guides gradient-based optimization to iteratively improve the NN parameters.

A similar training workflow has been explored in previous studies by Harp et al.~\cite{harp2021feasibility}, Srinivasan et al.~\cite{srinivasan2021machine}, and Pachalieva et al.~\cite{pachalieva2022physics}. However, the approaches by Harp and Srinivasan relied on simplified, differentiable analytical solutions, while Pachalieva et al. focused exclusively on single-phase steady-state flow. In contrast, our work is the first to integrate a fully differentiable transient multiphase flow simulator as part of the training loop, enabling more realistic and physically comprehensive modeling of subsurface pressure dynamics.

We employ a transfer learning strategy in our training workflow to mitigate the high computational cost associated with multiphase flow simulations. In the context of deep learning, transfer learning refers to a staged training approach in which the model first learns based on simpler tasks or data before being exposed to more complex scenarios. This approach is inspired by the way humans build knowledge, starting form simpler concepts and gradually progressing towards more complex ideas. This allows the model to fist capture generalizable patterns and then fine-tune it's predictions using more challenging scenarios~\cite{bengio2009curriculum}.

In our transfer learning workflow, we first pre-train the CNN using a single-phase steady-state flow model to accelerate convergence and reduce computational cost. The simulation setup for this stage is illustrated in Figures~\ref{fig:Application_pressure_setup}. The primary objective during pretraining is to enable the model to learn the mapping from heterogeneous permeability fields to extraction rates that maintain the prescribed pressure at a critical location, using the simplified single-phase steady-state formulation. Once the model achieves satisfactory accuracy in this setting, we transition to the more complex multiphase flow regime by replacing the single-phase simulator with the transient multiphase flow model. This sequential training strategy is illustrated in Figure~\ref{fig:Application_pressure_workflow_curriculum}, where the left panel represents the pretraining phase using steady-state single-phase flow, and the right panel depicts the final training phase involving time-dependent multiphase flow.

In both the single-phase and multiphase settings, the NN model predicts the extraction rate based on the input permeability field. Due to the consistency in simulation setup and input data structure across both phases and the high-level physical similarity, the transition from the single-phase to the multiphase model does not degrade the model’s predictive performance too badly. Additionally, since the pressure error is used as the training loss, which does not differ drastically between steady-state and transient simulations, the model is able to maintain its accuracy during the transition.

However, it is important to note that this transferability may not extend as well to other flow variables such as saturation in a multiphase flow. In such cases, the simplifications inherent in single-phase pretraining may not be sufficient to enable predictive accuracy during the transition to a multiphase framework.

\section*{Results}
In this section, we first present training results where we trained the NN from scratch using the transient multiphase flow solver to demonstrate the challenges of training directly on complex physics. Next, we show the training and validation losses obtained using the transfer learning workflow, which reflect the model's ability to manage the pressure. Then, we demonstrate the model’s effectiveness through a case study to confirm its ability to control pressure at the critical location. Finally, we evaluate the model's  performance on a set of randomly generated permeability fields, comparing the predicted extraction rates and the resulting pressure outcomes.

Before applying the transfer learning strategy, we trained the NN from scratch using data generated by the transient multiphase flow solver. This was done to illustrate the difficulties associated with learning complex physical dynamics without prior knowledge.
This baseline workflow mirrors the fine-tuning stage depicted in Figure \ref{fig:Application_pressure_workflow_curriculum}, with the key difference that the NN is initialized with random weights. To ensure that the model can capture meaningful variations in the pressure field, we simulated one year of fluid flow. The simulated pressure at the end of the run was compared to the target pressure, and the resulting difference was used to compute the training loss.

\begin{figure}[h!]
\centering
\includegraphics[width=0.6\linewidth]{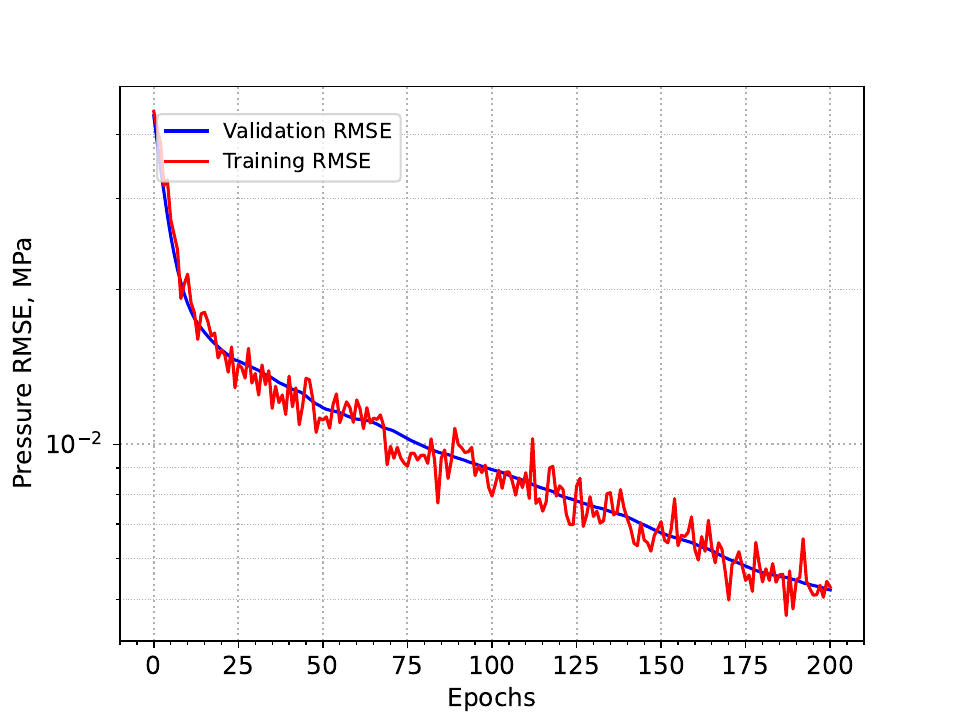}
\caption{Training and validation RMSEs from the initial training using only the multiphase solver. The high computational cost of incorporating complex physics makes this approach challenging, highlighting the need for more efficient learning strategies.}
\label{fig:Application_pressure_training_intial}
\end{figure}

The baseline training results confirm that the training NN models using multiphase flow simulations over extended run-time is computationally expensive. As shown in Figure~\ref{fig:Application_pressure_training_intial}, while the model begins to learn meaningful patterns, the convergence remains notably slow. To accelerate the process, we parallelized the training using Julia’s MPI framework, employing 40 processors in parallel. Despite this optimization, simulating one-year injection scenario over 200 training epochs required approximately 11 CPU hours with an AMD EPYC 7702P 64-Core processor. 
Later in this section, we show that applying transfer learning significantly improves model accuracy, while reducing the computational cost by an order of magnitude. These results highlight the inefficiency of relying solely on multiphase simulations over extended run-time during the early stages of training.

\begin{figure}[h!]
    \centering
    \includegraphics[width=0.6\textwidth]{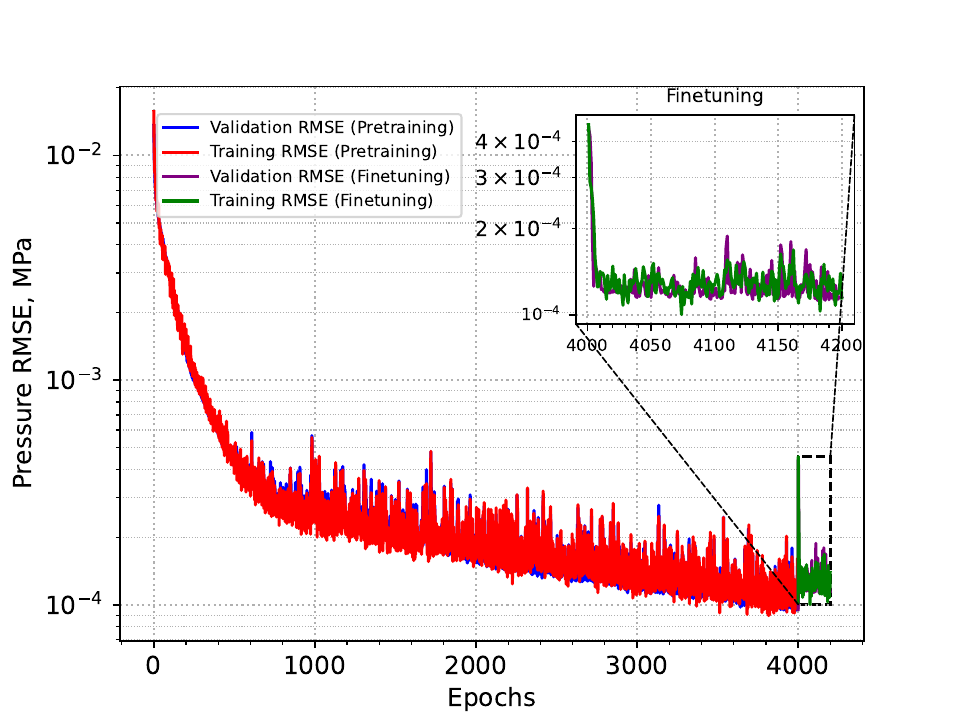}
    \caption{Training and validation RMSEs over the course of transfer learning. The smooth convergence of both curves indicates a stable knowledge transfer, with error approaching to a non significant value. By the end of training, the pressure error falls to as low as 0.0001\,MPa.}
    \label{fig:Application_pressure_overall_training}
\end{figure}

Next, we used our transfer learning workflow to train NN model, which determine the extraction rate at the extraction well.  Figure~\ref{fig:Application_pressure_overall_training} presents the RMSE observed during both the training and validation stages of the transfer learning workflow. The red and blue lines correspond to the validation and training errors during the pretraining phase, respectively, while the yellow and green lines represent the validation and training errors during the final multiphase training phase.
As illustrated in the figure, transitioning from the single-phase steady-state simulator to the transient multiphase simulator only marginally degrades the model’s performance. Both training and validation errors continue to decline steadily throughout the training process, indicating a smooth and effective transfer of learned representations between physical models. A closer examination of the zoomed-in final training results reveals that the model achieves its minimum error within just 20 epochs after the transition.

The total training time was approximately 5 CPU hours, including 1 hour for the pretraining phase and 4 hours for finetuning with the multiphase flow solver -- highlighting a significant gain in training efficiency. Most of the loss reduction during fine-tuning occurs within the first 13 epochs, requiring only about 0.3\,CPU hours. If training is halted at that point, the total training time drops to just 1.3\,CPU hours. The results indicate that incorporating transfer learning allows the model to maintain high accuracy while reducing overall computational cost by an order of magnitude.

As shown by the training and validation errors in Figure~\ref{fig:Application_pressure_overall_training}, the pretrained model rapidly reaches a high level of accuracy, with errors dropping well below 0.0001 MPa. This level of precision indicates that the model has undergone sufficient training to accurately predict optimized extraction rates for a wide range of heterogeneous permeability fields. To further test this, we evaluate the model’s performance on a set of randomly generated permeability realizations and analyze the resulting pressure responses using the predicted extraction rates. The testing procedure is described in detail in the following evaluation section.

\subsection*{Evaluation of pressure management workflow}
After training the NN model, we employ it to predict suitable extraction rates and subsequently calculate the resulting overpressure at the critical location using a full physics multiphase simulation. This process allows us to evaluate the effectiveness of the NN model in accurately determining extraction rates to maintain the desired pressure near the critical location. 

\begin{figure}[h!]
    \centering
    \begin{subfigure}[b]{0.49\textwidth}
        \includegraphics[width=\textwidth]{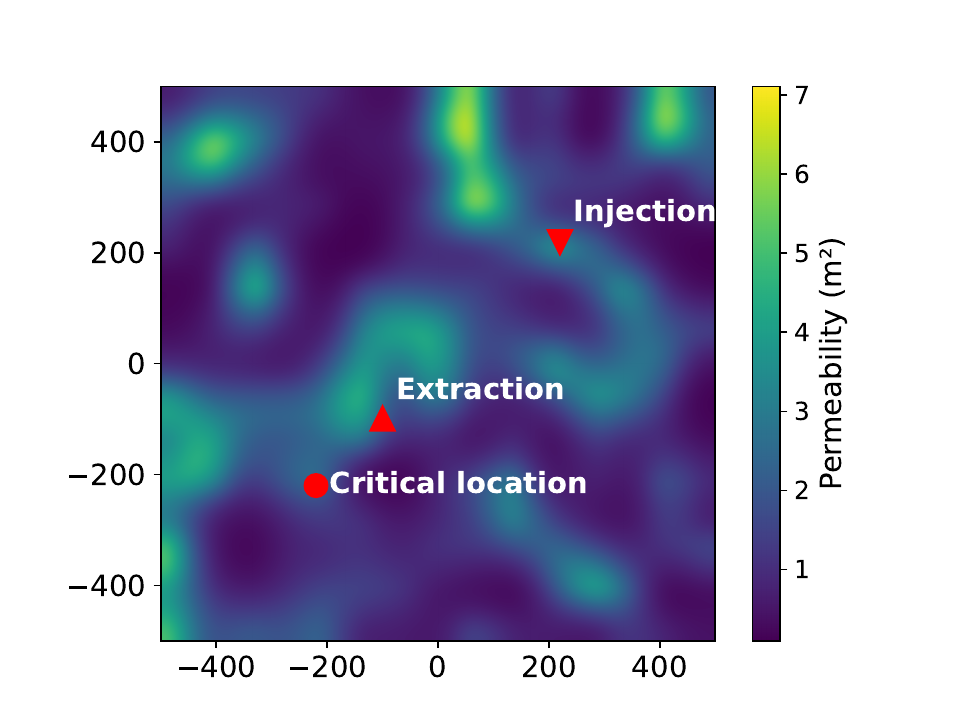}
        \caption{}
        \label{fig:Application_pressure_test_perm}
    \end{subfigure}
    \hfill 
    \begin{subfigure}[b]{0.49\textwidth}
        \includegraphics[width=\textwidth]{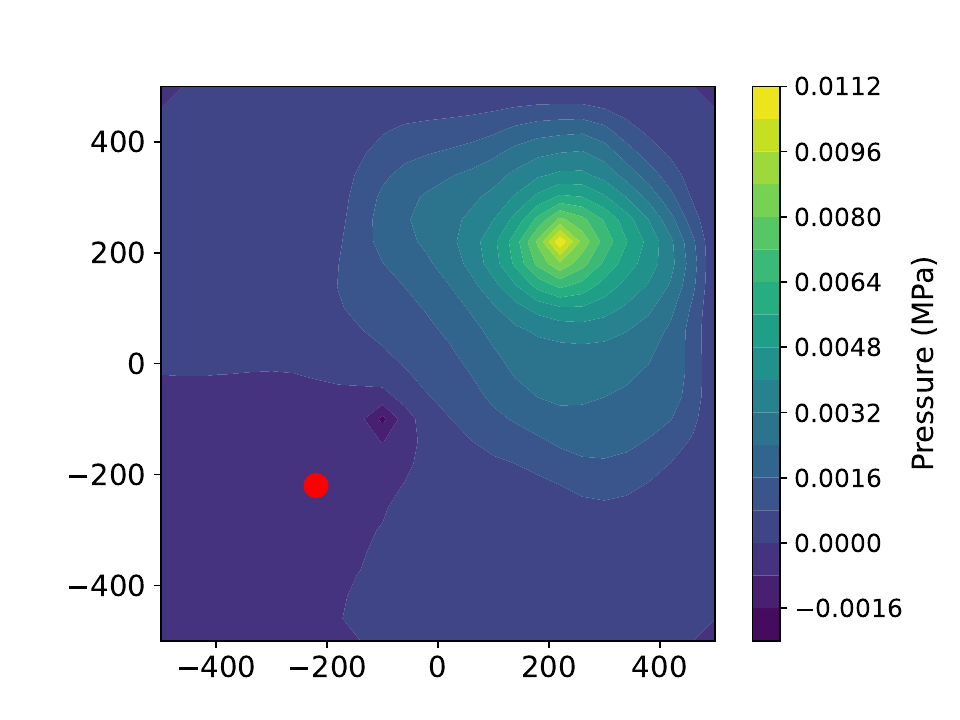}
        \caption{}
        \label{fig:Application_pressure_test_contour}
    \end{subfigure}
    \caption{Evaluation of the trained model. (a) Permeability field and well locations used for the test prediction case; (b), corresponding pressure contour from the simulation, where the red dot indicates the critical location.This single evaluation with a random permeability field shows that the model can successfully predict a suitable extraction rate which maintain zero pressure at the critical location.}
    \label{fig:Application_pressure_overall_training_eval}
\end{figure}

Initially, we demonstrate this approach using a sample permeability field, illustrated in Figure \ref{fig:Application_pressure_test_perm}. The corresponding pressure field generated by full physics simulation is shown in Figure \ref{fig:Application_pressure_test_contour}. In this figure, a negative pressure region is clearly observed around the extraction well, indicative of fluid withdrawal. Moreover, the pressure reaches zero around the critical location, which is highlighted by a red dot in the figure, matching the prescribed pressure for this location. This result confirms the NN model's capability to successfully predict an extraction rate that maintains the specified pressure at a critical reservoir location.

\begin{figure}[h!]
    \centering
    \begin{subfigure}[b]{0.49\textwidth}
        \includegraphics[width=\textwidth]{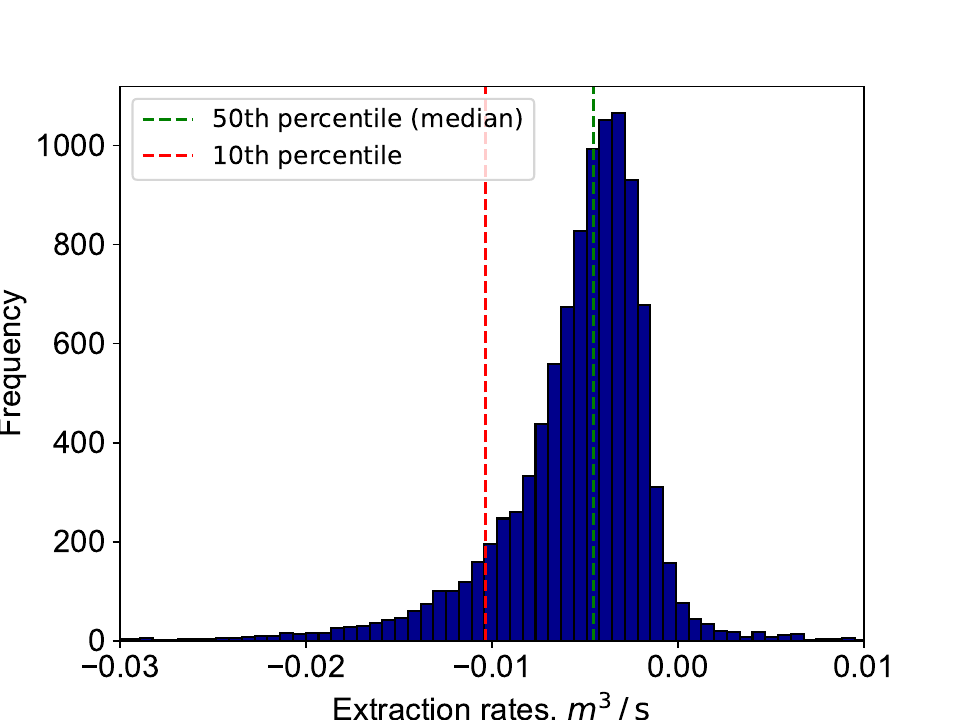}
        \caption{}
        \label{fig:Application_pressure_test_extractionRate}
    \end{subfigure}
    \hfill 
    \begin{subfigure}[b]{0.49\textwidth}
        \includegraphics[width=\textwidth]{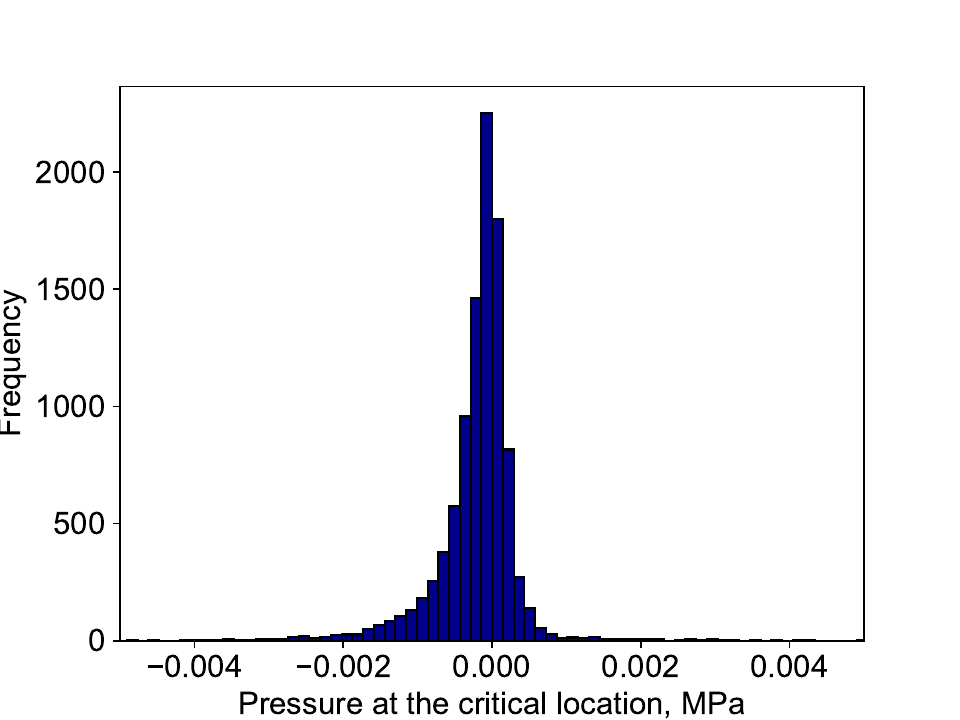}
        \caption{}\label{fig:Application_pressure_test_extractionPressure}
    \end{subfigure}
    \caption{(a) Extraction rates in cubic meters per second [$m^3/s$]; (b) resulting pressure at the critical location in mega pascals [$MPa$]. For input of 10,000 random permeability fields the NN model predicts extraction rates, which maintains  zero or near zero pressure at the critical location for most the cases. It confirms that the trained model can be used to handle uncertainty and heterogeneity for pressure management in a field.}
    \label{fig:Application_pressure_test}
\end{figure}

Next, we analyze the distribution of the predicted extraction rates and the corresponding overpressure at the critical location. To generate these distributions, we pass 10,000 randomly generated heterogeneous permeability fields (training set from Table~\ref{tab:param_NN}) through our trained NN model. For each case, we compute the resulting overpressure using the full physics simulation. The outcomes are presented in Figure \ref{fig:Application_pressure_test}, where the predicted extraction rates are shown in Figure \ref{fig:Application_pressure_test_extractionRate}, and the corresponding overpressure values are illustrated in Figure \ref{fig:Application_pressure_test_extractionPressure}. The distribution of extraction rates is notably skewed, with 90\% of the samples requiring an extraction rate below 0.01\,m\textsuperscript{3}/s (equivalent to 0.32\,Mt/year) to maintain the target pressure. The average extraction rate is 0.0046\,m\textsuperscript{3}/s, or 0.15\,Mt/year. Compared to the injection rate of approx.\,1\,Mt/year, the average extraction rate is about 15\%, which means for 50\% of samples we need to extract 15\% or less of the injected fluid. Meanwhile, the overpressure distribution reveals only minor deviations from the prescribed pressure of zero at the critical location for most of the samples. 

These results ensure the model's capability to consistently predict extraction rates that maintain the target pressure at critical points within the reservoirs with random heterogeneous permeability field. The pressure evaluation shows improved accuracy compared to past work that handled only single phase flow~\cite{pachalieva2022physics}. This makes the framework highly applicable for real-field pressure management tasks, enabling rapid, data-driven control decisions in complex subsurface environments.

\section*{Discussion}
In this study, we demonstrated that our pressure management workflow can accurately and efficiently manage pressure in heterogeneous subsurface reservoirs by controlling the fluid extraction rate. Similar approaches have previously been developed using simpler physics models, such as homogeneous permeability~\cite{harp2021feasibility} and single-phase flow~\cite{pachalieva2022physics}. We incorporate more complex physics into our model to represent a more realistic approximation of field operations. Our workflow uses a heterogeneous permeability field and transient multiphase flow to improve prediction accuracy, while the use of automatic differentiation within the DPFEHM framework, and transfer learning enhances the efficiency of the model. This efficient and physics‑based approach to pressure management will enable effective control of reservoir pressure and will aid in real‑time decision making during injection and extraction operations in many subsurface applications, including gas storage, oil and gas production, wastewater disposal, and geothermal energy extraction~\cite{ben2006info,charnes1978measuring,geng2001intelligent,zimmermann2009pressure, zhang2013decision,simmenes2013importance, zhou2024managing}.

The advantages of our workflow are as follows:
\begin{itemize}
    \item Our model successfully predicts the extraction rate and maintains prescribed pressure at critical well locations.
    \item Due to the automatic differentiability in our background physics model, we are able to use a large number of ML parameters for training and testing.
    \item Our workflow uses a CNN model, which is fast and produces small errors (less than 1\,kPa).
    \item We create the training and testing datasets on the fly using the fully differentiable physics model and random realizations, rather than storing and loading data like many ML approaches.
    \item The large training data makes the model suitable for handling uncertainty in formation properties.
    \item Our workflow successfully uses a transfer learning to reduce the high computational cost associated with complex transient and multiphase flow by several orders of magnitude, which was identified as a major bottleneck in past work~\cite{harp2021feasibility}.
    \item By using transient multiphase flow, our model covers a wider range of applications than past work that used single‑phase, steady‑state flow and homogeneous properties~\cite{harp2021feasibility, pachalieva2022physics}.
\end{itemize}

Although the presented workflow brings significant improvements over previous approaches and more accurately represents real‑world complex physics, it still has some limitations and scope for improvement. Which are listed as follows:

\begin{itemize}
    \item The current workflow is based on a simplified two-dimensionalreservoir geometry. Future work should extend this to three-dimenional models, which more accurately represent real-world CO$_2$ storage sites. 
    \item The impact of fluid properties on the learning process remain unexplored. Future studies should investigate how variations in fluid characteristics influence the effectiveness and transferability of the trained models.
    \item The background physics model was run for one year; however, gas storage operations typically span much longer timescales. A future study to extend the workflow should include simulation periods on the scale of years to more accurately represent field‑scale pressure management.
\end{itemize}

\section*{Conclusions}
We have introduced a novel pressure management workflow that integrates a fully differentiable, transient multiphase reservoir simulator with a CNN. Unlike previous work on that rely on homogeneous properties or steady-state assumptions, our approach incorporates heterogeneous permeability fields and time-dependent multiphase flow, enabling high-fidelity pressure control at critical well locations.

One of the key strengths of this workflow is that by employing transfer learning from steady-state single-phase simulations to more complex multiphase simulations, we dramatically reduce computational cost while producing a highly accurate model. Another advantage of our workflow is its ability to generate large number of random training and testing data on the fly to effectively capture subsurface uncertainty. The use of automatic differentiation within the DPFEHM framework allows us to integrate of complex physics into the training process. 

When evaluated across a wide range of randomly generated permeability field, the workflow consistently predicts near-optimal extraction rates and maintains target pressures. The proposed approach delivers results that are not only accurate but also significantly faster than those produced by full-physics simulators at inference time, making it well-suited for near real-time applications. This workflow represents a meaningful step toward robust, efficient, and accurate pressure management across a wide range of real-world reservoir operations.
\section*{Data availability}
The datasets generated and/or analysed during the current study are available in the [DPFEHM] repository, [\href{https://github.com/OrchardLANL/DPFEHM.jl}{DPFEHM.jl}]
\bibliography{main}

\section*{Funding}
This work is funded by the U.S. Department of Energy, Office of Science Energy Earthshot Initiative as part of the project ``Learning reduced models under extreme data conditions
for design and rapid decision-making in complex systems" under Award number no. DE-SC0024721. DO was partially funded by U.S. Department of Energy, Office of Science (Basic Energy Sciences) Early Career Award number ECA1.
\section*{Acknowledgments}

This material is based upon work supported by the U.S. Department of Energy, Office of Science Energy Earthshot Initiative as part of the project ``Learning reduced models under extreme data conditions
for design and rapid decision-making in complex systems" under Award number no. DE-SC0024721. DO was partially supported by U.S. Department of Energy, Office of Science (Basic Energy Sciences) Early Career Award number ECA1.

\section*{Author contributions statement}
H.R.: Data curation, Formal analysis, Investigation, Methodology, Software, Validation, Visualization, Writing -- original draft, Writing -- review and editing
A.P.: Methodology, Software, Writing -- review and editing.
D.O.: Conceptualization, Formal analysis, Software, Methodology, Writing -- review and editing.


\end{document}